\documentclass[letterpaper, 10 pt, conference]{ieeeconf}  

\IEEEoverridecommandlockouts                              

\usepackage{times}
\usepackage{epsfig}
\usepackage{graphicx}
\usepackage{amssymb}
\usepackage{amsmath}
\usepackage[linesnumbered,ruled]{algorithm2e}
\usepackage{array}
\usepackage{rotating}
\usepackage{multirow}
\usepackage{csquotes}
\usepackage{color}
\newcolumntype{P}[1]{>{\centering\arraybackslash}p{#1}}

\newcommand{\thickhline}{%
    \noalign {\ifnum 0=`}\fi \hrule height 1pt
    \futurelet \reserved@a \@xhline
}

\usepackage{subcaption}
\usepackage{graphicx}

\title{\LARGE \bf
Domain and View-point Agnostic Hand Action Recognition
}

\author{Alberto Sabater$^{1}$ \hspace{0.5cm} Iñigo Alonso $^{1}$ \hspace{0.5cm} Luis Montesano$^{1,2}$ \hspace{0.5cm} Ana C.~Murillo$^{1}$

\thanks{This research has been funded by 
PGC2018-098817-A-I00 (MCIU/AEI/FEDER, UE), DGA T45 17R/FSE and the Office of Naval Research Global project ONRG-NICOP-N62909-19-1-2027.}

\thanks{$^{1}$ A. Sabater, I. Alonso, L. Montesano and A.C. Murillo are with 
DIIS - I3A, Universidad de Zaragoza, Spain. {\tt\small \{asabater, ialonso, montesano, acm\}@unizar.es}}
\thanks{$^{2}$  L. Montesano is also with Bitbrain Technologies, Zaragoza, Spain. {\tt\small \{luis.montesano\}@bitbrain.com}}
}

\begin{document}

\maketitle
\thispagestyle{empty}
\pagestyle{empty}

\begin{abstract}
Hand action recognition is a special case of action recognition with applications in human-robot interaction, virtual reality or life-logging systems. Building action classifiers able to work for such heterogeneous action domains is very challenging. There are very subtle changes across different actions from a given application but also large variations across domains (e.g. virtual reality vs life-logging). This work introduces a novel skeleton-based hand motion representation model that tackles this problem. The framework we propose is agnostic to the application domain or camera recording view-point. When working on a single domain (intra-domain action classification) our approach performs better or similar to current state-of-the-art methods on well-known hand action recognition benchmarks. And, more importantly, when performing hand action recognition for action domains and camera perspectives which our approach has not been trained for (cross-domain action classification), our proposed framework achieves comparable performance to intra-domain state-of-the-art methods. These experiments show the robustness and generalization capabilities of our framework.
%

\end{abstract}


\section{Introduction}
Human action recognition is a well studied problem with many applications such as human-robot interaction, surveillance and monitoring \cite{krupke2018comparison, tanwani2017generative}. Deep models combined with skeleton-based representations \cite{moon2018v2v, xiong2019a2j}, which efficiently encode human pose and motion, independently of appearance, surroundings and occlusions, have become a standard in robust human action recognition \cite{perez2019interaction, zhang2019view}.

Hand action recognition is a specific case of human action recognition. It is highly relevant due to the importance of hand movements in team work, assistive technologies, communication or in virtual reality applications \cite{abbasi2019multimodal, bates2017line}. 
Hand recognition methods combine, as in the full body case, skeleton representations and deep models \cite{yang2019make, zhang2016efficient}. These methods have shown good results typically focusing  on the classification of actions from a specific domain.
However, previous works have not studied robust representations that can generalize across different domains and view-points, that are key when working with limited amounts of labeled data.  

Hand actions expose some specific challenges to learn such robust representations. 
On one hand, there is a high variability across actions from different domains, e.g. user interface control vs. life-logging applications. Differently from full-body skeletons, different hand action domains often imply drastic view-point changes, e.g. egocentric vs. third-person view. 
On the other hand, fine grained details are essential. Different action categories are often quite similar and vary only subtly (e.g. pointing to different directions, sliding gestures, etc.). 
Moreover, hand skeleton joints present lower movement range than other full-body joints, increasing the correlation of skeleton joint motions and similar actions.

\begin{figure}[!tb]
    \centering
    \includegraphics[width=0.9\columnwidth]{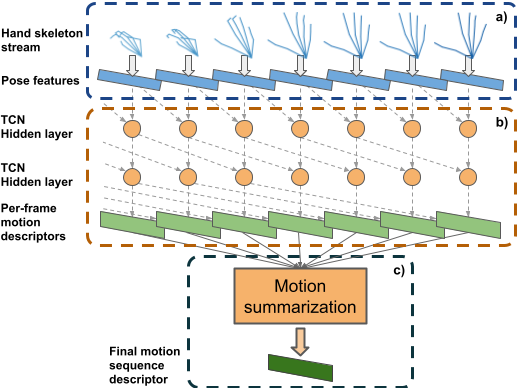}
    \caption{Motion representation model.
    \textbf{(a) Hand pose modeling}: 
    pose features are extracted from the input hand skeleton stream.
    \textbf{(b) Motion representation}: 
    a Temporal Convolutional Network (TCN) generates per-frame motion descriptors from the pose features.
    \textbf{(c) Motion summarization}: 
    per-frame motion descriptors are weighted and summarized to generate the final motion sequence descriptor.
    }
    \label{fig:pipeline}
\end{figure}

The main contribution of this work\footnote{Code, learned models, supplementary video and data splits can be found in: https://sites.google.com/a/unizar.es/filovi/} is a novel motion representation model, summarized in Fig. \ref{fig:pipeline}, designed to be robust to different application domains and view-points. It computes representations (motion descriptors) from labeled hand skeletons (motion sequences), that are later used for the final motion sequence classification.
The main components of our 
motion representation model are: 1) a set of pose features adapted to hand motion; 2) a Temporal Convolutional Network (TCN) encoding the stream of hand pose features into per-frame descriptors; and 3) a summarization module that learns the relevance of each per-frame descriptor to describe the input motion sequence.  
The learned motion descriptors can be directly used to recognize the action categories (labels) they were trained for (intra-domain) with a simple Linear Classifier. More interestingly, they can also be used to build N-shot classifiers to recognize new unseen action categories recorded from radically different points of view (cross-domain) with a K-Nearest Neighbor Classifier.

Our experiments use the front view SHREC-17 dataset~\cite{de2017shrec}, the egocentric F-PHAB dataset \cite{garcia2018first} and the third-person MSRA \cite{sun2015cascaded} dataset, which include actions and gestures related to computer interaction, life-logging and sign language domains respectively.
Our intra-domain classification results show that our framework gets better or similar performance than current state-of-the-art intra-domain classifiers in well-known benchmarks.
More importantly, our cross-domain classification approach obtains comparable accuracy to intra-domain methods by being trained just with the SHREC-17 dataset, and then evaluated on the F-PHAB and MSRA datasets. This demonstrates that our motion representation model generalizes well for different action domains and camera view-points.
Besides, our approach shows a low latency, which allows its use for online and real-time applications.

\section{Related Work}

This section summarizes relevant works on the core topics of this work: pose modeling, skeleton-based action recognition models and generalization to unseen action categories.

\subsection{Pose modeling for action recognition}
Action recognition was first tackled by directly analyzing RGB videos \cite{feichtenhofer2016convolutional} or depth maps \cite{oreifej2013hon4d}. Current approaches have settled the standard of extracting the intermediate representation of skeleton poses \cite{zhang2016efficient, yang2019make}. This representation has shown great performance since it encodes human poses regardless their appearance and surrounding and presents strong robustness to occlusions.

Certain works directly use the raw coordinates of skeleton joints (position of the joints in the Euclidean space) as input for full-body action recognition \cite{perez2019interaction, liu2019ntu} and for hand action recognition \cite{hou2018spatial, ma2020skeleton, li2021two}. 
In order to achieve a standardized and generic skeleton pose descriptions, several full-body action recognition approaches propose different strategies, such as learning the most suitable view-point for each action \cite{zhang2019view} or transforming all coordinates to a common coordinate system \cite{sabater2021oneshot, su2020predict}. However, this kind of transformations cannot be directly applied to hand action recognition, where orientation plays a key role. 

In order to get more informative pose representations than the raw joint coordinates, many approaches propose to compute additional geometric (pose) features. Chen et al. \cite{chen2010learning} use static features (distance and angles of pairs of joint coordinates) and temporal features (velocity and acceleration of joint coordinates).
Zhang et al. \cite{zhang2017geometric} calculate distances between joints and planes, and Yang et al. \cite{yang2019make} use joint distances and their motion speeds at different scales.

Our approach proposes a simplification of the skeleton representation reducing coordinate redundancy by using just a set of key joints. 
Then, simplified skeleton coordinates are standardized by applying scale and location invariant transformations. Specific geometric features are calculated to encode relevant translation information (lost in the standardization) and orientation aware information.

\subsection{Action recognition models}

As in many other fields, deep learning has become state-of-the-art in action recognition. Particularly relevant for this work, Recurrent Neural Networks (RNN) have been widely used to model temporal dependencies in hand action recognition.
Ma et al. \cite{ma2020skeleton} use a LSTM-based Memory Augmented Neural Network to model dynamic hand gestures. Chen et al. \cite{chen2019mfa} use a LSTM Network to combine skeleton coordinates, global motions and finger motion features. Li et al.  \cite{li2021two} combine a bidirectional Independently Recurrent Neural Network with a self-attention based graph convolutional network.

Other works make use of Convolutional Networks. Liu et al. \cite{liu20203d} recognize posture and action by using 3D convolutions. Yang et al. \cite{yang2019make} use  1D convolutions to process and fuse different hand motion features. Hou \cite{hou2018spatial} propose  to focus on the most informative hand gesture features by using a  ResNet-like 1D convolutional network with attention.

Our method uses a Temporal Convolutional Network (TCN) \cite{bai2018empirical, oord2016wavenet} that implements 1D dilated convolutions to learn long-term temporal dependencies from variable-length input sequences, achieving comparable or better results than RNNs \cite{bai2018empirical}. 
TCNs have demonstrated good performance on full-body action recognition, both with unsupervised learning \cite{su2020predict} and supervised learning \cite{sabater2021oneshot, kim2017interpretable}.

\subsection{Generalization to unseen action categories}

Learning a model able to classify unseen categories is a challenging task. It is commonly tackled by encoding every new data sample into a descriptor and using a K-Nearest Neighbors classifier (KNN) to evaluate and assign labels according to the similarity between a few new category reference samples and the target samples \cite{wang2019simpleshot}.

Several works \cite{liu2019ntu,sabater2021oneshot} address this problem for action recognition by extracting intermediate feature maps from a supervised action recognition model.
Koneripalli et al. \cite{koneripalli2020rate} train an autoencoder to learn these descriptors in an unsupervised fashion. Ma et al. 
\cite{ma2020skeleton} learn these descriptors directly in a semi-supervised manner by training an encoder with metric-learning techniques.
Other works use word2vec \cite{hahn2019action2vec} and sent2vec \cite{jasani2019skeleton} approaches for descriptor learning.

Previous works \cite{liu2019ntu,hahn2019action2vec,jasani2019skeleton} are aimed to recognize unseen full-body action categories where no drastic camera view-points are found. Up to our knowledge, generalization to unseen hand view-points and domains is still to be studied.
The present work uses metric-learning and specific data augmentation to learn meaningful hand sequence descriptors. Our framework performs accurate action recognition of sequences from unseen categories and recording view-points.

\section{Hand action recognition framework}

The core of the proposed framework is the motion representation model summarized in Fig. \ref{fig:pipeline}. 
First, our approach calculates specific pose features for each skeleton (in our case already pre-computed and available in common datasets, see Section~\ref{sec:datasets}). These features are fed to a Temporal Convolutional Network to generate a set of motion descriptors.
Additionally, a motion summarization module combines them, according to their relevance, into the final motion representation.
In the following, we describe these steps, as well as how to train our motion representation model, both for intra-domain and cross-domain classification.

\subsection{Hand pose modeling}\label{sec:hand_modeling}

Human hand motion sequences are defined by sets of \(T\) hand skeleton poses \(X=\{X_1, ..., X_T\}\), extracted from video frames. 
Each hand skeleton \(X_t\) is composed by a set of \(J\) joint coordinates, $X_t = \{x_1, ..., x_J \}, x_j \epsilon \mathbb{R}^3$ (i.e. position of the joints in Euclidean space), which are logically connected by a set of \(B\) bones 
(see Fig. \ref{fig:min_hand}).

\subsubsection{Skeleton standardization}
Since motion information has high variability across different action domains, we propose several steps to standardize the skeleton representation to help generalization of the motion representation model.

\begin{figure}
    \centering
    \includegraphics[width=0.70\linewidth]{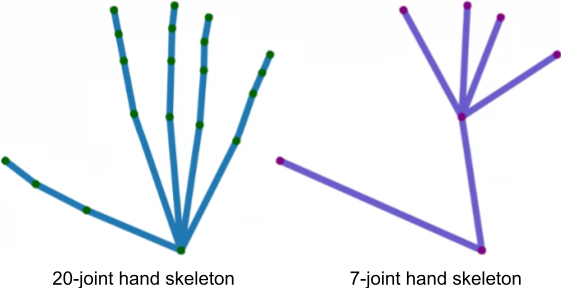}
    \caption{Hand skeleton simplification. Left diagram refers to a detailed hand skeleton of 20 joints (dots) connected by 19 bones (lines). Right diagram refers to our proposed hand skeleton simplification of 7 joints (dots) and 5 bones (lines).
    }
    \label{fig:min_hand}
\end{figure}

First, hand joints belonging to the same bones (fingers) are highly coupled and can be represented with a smaller number of degrees of freedom.
Based on this assumption, we propose to use just a subset of \textbf{7 joints to define a hand pose} (see Fig. \ref{fig:min_hand}), corresponding to the wrist, the top of the palm, and the tips of the 5 fingers; which we connect with a total of \textbf{6 hand bones}, one for the palm and one more for each one of the fingers. 
This simpler skeleton representation makes the learning process easier and less prone to overfitting.

Secondly, since actions can be performed by different people with heterogeneous hand sizes and recorded at different scales, we standardize each skeleton pose \(X_t\) to achieve \textbf{scale-invariant} skeleton representations \(\hat{X}_t\) by applying, to all the hand coordinates, the transformation that makes the palm of size equal to 1: 

\begin{equation}
\hat{X}_{n}=\frac{{X}_{n}}{\left | P \right |},
\end{equation}
where \(\left | P \right |\) is the euclidean distance between the wrist and the top of the palm  (both joints included in original and simplified 7-joint formats).

Finally, since actions must be recognized regardless the position where they are executed, we compute \textbf{location-invariant} coordinates (relative coordinates) by translating the top of the palm to the origin of the reference coordinate system. Note that these relative hand coordinates describe properly the intra-relation of the hand joints, but they are now missing the information related to the hand motion direction.

\subsubsection{Hand pose description}

Different from full body motion sequences (e.g. walking) where their movement direction can be inferred from the relative coordinates of its bones (e.g. legs), hands can be translated through any direction without any change of their relative coordinates. Since the translation information is essential in certain actions (e.g. pointing to specific directions), we generate extra \textbf{translation and orientation-aware} features from the original hand skeletons:
\begin{itemize}
    \item \textbf{Difference of coordinates}, defined as the difference of each joint coordinate with itself in the previous time-step. These features describe the translation direction and speed of each coordinate for each of the 3 axes:
        \begin{equation}
            d_{coord}(t,j) = x_{j,t} - x_{j,t-1}, \forall j \epsilon J, t \epsilon T
        \end{equation}
    \item \textbf{Difference of bone angles}, defined as the difference of the elevation \(\varphi\) and azimuth \(\theta\) of a bone \(b \epsilon B\) with itself in the previous time-step. These features describe the rotation direction (with respect to the world coordinates) and rotation speed of each bone:
        \begin{equation}
            d_{\varphi}(t, b_\varphi) = b_{\varphi,t} - b_{\varphi,t-1}, \forall b \epsilon B, t \epsilon T
        \end{equation}
        \begin{equation}
            d_{\theta}(t, b_\theta) = b_{\theta,t} - b_{\theta,t-1}, \forall b \epsilon B, t \epsilon T
        \end{equation}
\end{itemize}

Our final hand representation is a feature vector of size \(54\) (Fig. \ref{fig:pipeline}.a) ($7\times3$ relative hand coordinates, $7\times3$ coordinate difference features, and $6\times2$ bone angle differences).

\subsection{Motion representation model}\label{sec:tcn}

The core of our action recognition framework is a model that encodes the skeleton features from each frame, described in the previous section, into single motion descriptors with a Temporal Convolutional Network (TCN) \cite{bai2018empirical, oord2016wavenet}. 
The TCN processes sequences of skeleton features, generating a descriptor at each time-step per-frame descriptors) that represents the motion performed up to that frame, i.e. with no information from the future (see Fig. \ref{fig:pipeline}.b).

\begin{figure}
    \centering
    \includegraphics[width=0.48\textwidth]{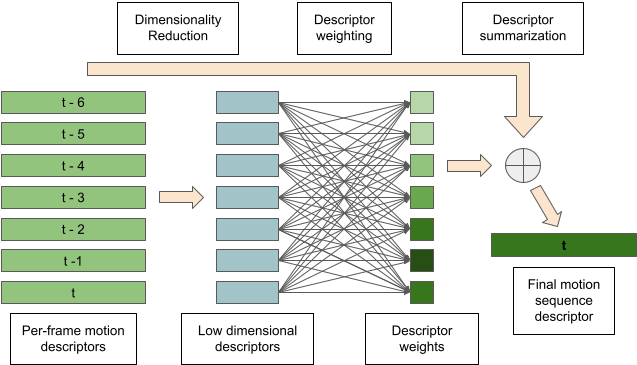}
    \caption{Motion summarization module. Per-frame motion descriptors are simplified with a 1D Convolutional layer. They are grouped with a single perceptron layer to calculate their summarization weights. The motion sequence is summarized into a final motion descriptor by performing a weighted average over the initial per-frame motion descriptors.}
    \label{fig:summarization}
\end{figure}

For a given motion sequence, the last descriptor generated by the TCN is frequently the one used to represent the action \cite{sabater2021oneshot}, since it encodes all the information up to that point. 
However, training motion sequences are not frequently segmented in time with high precision. In these cases, sequence endings contain frames that are not informative for the action they represent. Consequently, the last descriptor can introduce some noise that hinders the training.

To alleviate this issue, we learn the relevance of the temporal patterns of the actions. More precisely, we add a \textbf{motion summarization module} after the TCN (see Fig. \ref{fig:pipeline}.c), which combines all the per-frame descriptors generated for 
the input hand motion, 
up the TCN memory length, by performing a weighted average over them (details in Fig. \ref{fig:summarization}). 
These weights represent how important each descriptor is for the final motion representation. They are learned with a simple Neural Network trained end-to-end along with the TCN. 
This network consists of a single 1D Convolutional layer with kernel 1 that reduces the per-frame descriptors dimensionality, and a single Fully Connected layer with a sigmoid activation layer, that takes as input all the simplified descriptors and outputs a vector of categorical probabilities (i.e. descriptor weights). These final weights are L1 normalized before performing the final descriptor summarization. 

This summarization module efficiently describes hand motion sequences and helps the TCN to focus just on the meaningful data during training. However, there are real use cases where actions, at test time, present a longer length than our motion representation module can handle. In these cases, although the summarization module has been trained along with the TCN, it is better to discard it and classify individually all the per-frame descriptors generated by the TCN, which still contain meaningful motion representations.

So far, we have shown how to encode a motion sequence $X$ into a robust simple descriptor $z = f(X)$, where the function $f$ represents our motion representation module (Fig. \ref{fig:pipeline}). In the next two sections, we describe how to optimize these motion representations to perform intra-domain classification and cross-domain classification.

\subsection{Intra-domain classification}\label{sec:intra_dom}

Intra-domain hand action classification aims to recognize the same actions categories (labels) seen during the learning phase, with no drastic variation on the camera view-point. For this classification, intra-domain class probabilities $P = g(z)$ are predicted by a linear classifier \(g\) trained end-to-end along with our motion representation model \(f\) (represented in Fig.~\ref{fig:pipeline}). Intra-domain classification is learnt by the optimization of the categorical cross-entropy loss:

\begin{equation}
CCE = -\Sigma_{c=1}^{C} y_{i, c} \log \left(p_{i, c}\right),
\end{equation}
which evaluates the predicted probabilities \(p_{i,c}\) that belongs to a class \(c \epsilon C\), given their true label \(y_i\).

Each training iteration include a mini-batch composed of motion sequences sampled uniformly for each action category (2 different samples per category in our experiments).
To ensure the generalization to different motion artifacts, which can be hard to achieve with small datasets,
each motion sequence within the mini-batch is included three times with different data augmentations. This data augmentation is applied to the per-frame skeletons \(X_t\), before the feature computation from Section \ref{sec:hand_modeling}, as follows:
\begin{itemize}
    \item Movement speed variation. Joint coordinates are randomly re-sampled by interpolation over the temporal dimension. This simulates different motion speeds, and thus, different sequence lengths.
    \item Frame skipping. Since contiguous video frames contain similar joint information, we only use one out of every three frames, reducing the data redundancy and making the learning process easier. Motion sequences are then initialized randomly between the three first frames.
    \item Random cropping. When the sampled motion sequence is longer than a defined maximum length (i.e. TCN memory lenght), it is randomly cropped.
    \item Random noise. Gaussian noise is added to the skeleton coordinates to simulate inaccurate joint estimations.
    \item Random rotation noise. The whole motion sequence is rotated randomly over the 3D axes. This rotation is limited to low angles, to simulate just subtle variations in the recording view-point. 
\end{itemize}

\subsection{Cross-domain classification}\label{sec:xdom}

Cross-domain hand action classification aims to recognize motion sequences whose action category and recording camera view-point were not present in the training data. To obtain view-point agnostic motion representations, our motion representation model \(f\) is trained, via contrastive learning, to project motion descriptors in a space where
descriptors belonging to the same action category (label) must be close to each other (similar descriptors), and far away from other category descriptors (dissimilar descriptors).
This is achieved optimizing 
the \textit{normalized temperature-scaled cross-entropy loss} (\textbf{NT-Xent}) \cite{chen2020simple}:
\begin{equation}
    l_{i, j}=-\log \frac{\exp \left(\operatorname{sim}\left(z_{i}, z_{j}\right) / \tau\right)}{\sum_{k=1}^{2 N} 1_{[k \neq i]} \exp \left(\operatorname{sim}\left(z_{i}, z_{k}\right) / \tau\right)},
\label{eq:loss}
\end{equation}

\noindent which is computed in each training iteration for each pair of actions \(i\) and \(j\) that belong to the same action category. 
NT-Xent maximizes the cosine similarity \(sim\) of both motion descriptors \(z_i\) and \(z_j\) and minimizes their similarity to the descriptors related to different action categories \(k\). \(\tau\) is a temperature parameter.

The training of our motion representation model is performed with the same batch construction and data augmentation techniques described in Section \ref{sec:intra_dom}. Additionally, we add an extra data augmentation step that rotates randomly all the motion sequences of the mini-batch over the three axis. This batch augmentation simulates arbitrary camera recording perspectives, which is crucial to boost the performance achieved with the NT-Xent loss in different domains and camera view-points.

Once this generic motion representation model has been trained on a given source domain, we use a N-shot approach~\cite{wang2019simpleshot} and generate motion descriptors for a small set of N reference motion sequences (motion reference set) from a different target domain, with no specific training on the latter. To perform action classification in this new domain, 
we use a simple K-Nearest Neighbors classifier (KNN) to assign a label to new sequences depending on their descriptor distance to the descriptors from the motion reference set. To improve the performance of the KNN, 
we extend our motion reference set by applying the same data augmentation strategies described in Section \ref{sec:intra_dom}, and we compute descriptors for all the new augmented sequences.


\section{Experiments}
This section details the datasets used in the evaluation and our implementation details.
Then, we expose the main framework design choices and evaluate its performance for cross-domain and intra-domain action recognition.
Finally, we evaluate the time-performance of the presented approach.

\subsection{Experimental setup}
\subsubsection{Datasets}\label{sec:datasets} The presented approach has been validated on three different datasets (see frame samples in Fig. \ref{fig:frame_samples}), with different application domains and camera view-points.

\paragraph*{SHREC-17 \cite{de2017shrec}} contains motion sequences (22-joint hand skeletons) related to human-machine interaction domains recorded from a \textbf{frontal third-person view}. The data is categorized with two levels of granularity, presenting 14 and 28 actions categories respectively. The dataset contains 1960 motion sequences for training and 840 sequences for validation. 
Actions are performed by 28 different users.

\paragraph*{F-PHAB \cite{garcia2018first}} contains motion sequences (21-joint hand skeletons) recorded from an \textbf{egocentric view} related to kitchen, office and social scenarios, which involve the interaction with different objects. 
Actions have been performed by 6 different users and labeled with 45 action categories.
The dataset consists of 1175 motion 
sequences which are split into training and validation as stated by the authors \cite{garcia2018first}:
    \textbf{1:3, 1:1, 3:1} splits the motion sequences on different training:validation ratios (e.g. in the 1:3 split, 33\% of the data is used for training and the remaining 66\% is used for validation);
    \textbf{cross-person} 6-fold leave-one-out cross-validation, one fold for the each user motion sequences.
Only the original cross-subject and 1:1 splits are available, for the other two data partitions we create three random data folds to perform 3-fold cross-validation.

\paragraph*{MSRA \cite{sun2015cascaded}} contains motion sequences (17-joint hand skeletons) of 17 different American Sign Language gestures performed by 9 different users. Each gesture sequence has a length of 500 frames recorded from a \textbf{third-person view}.
For the classification of this data, we use the motion samples from the two first subjects as reference, leaving the remaining seven as the target samples, as suggested in \cite{liu20203d}. \\

\begin{figure}[!bt] \label{fig:frame_samples}
\centering
   
\begin{subfigure}[b]{\linewidth} 
\centering
    \includegraphics[width=0.3\linewidth]{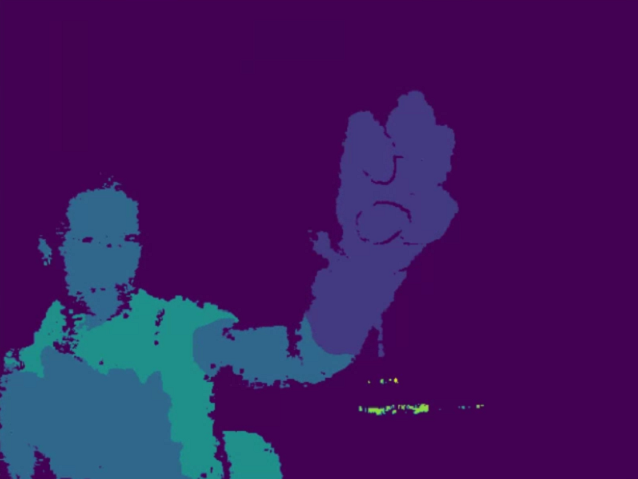}
    \includegraphics[width=0.3\linewidth]{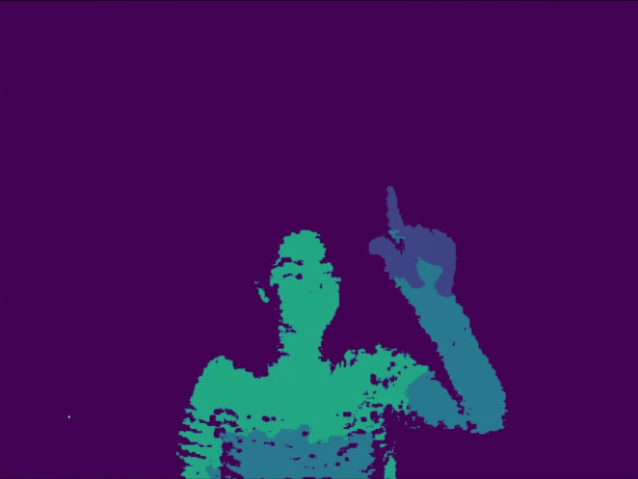}
    \includegraphics[width=0.3\linewidth]{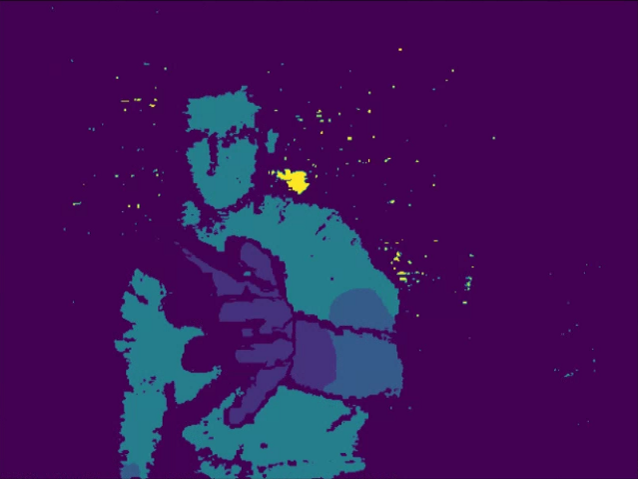}
    \caption{SHREC-17 depth sample frames}
\end{subfigure}
\begin{subfigure}[b]{\linewidth} 
\centering
    \includegraphics[width=0.3\linewidth, trim={0 15mm 0 0},clip]{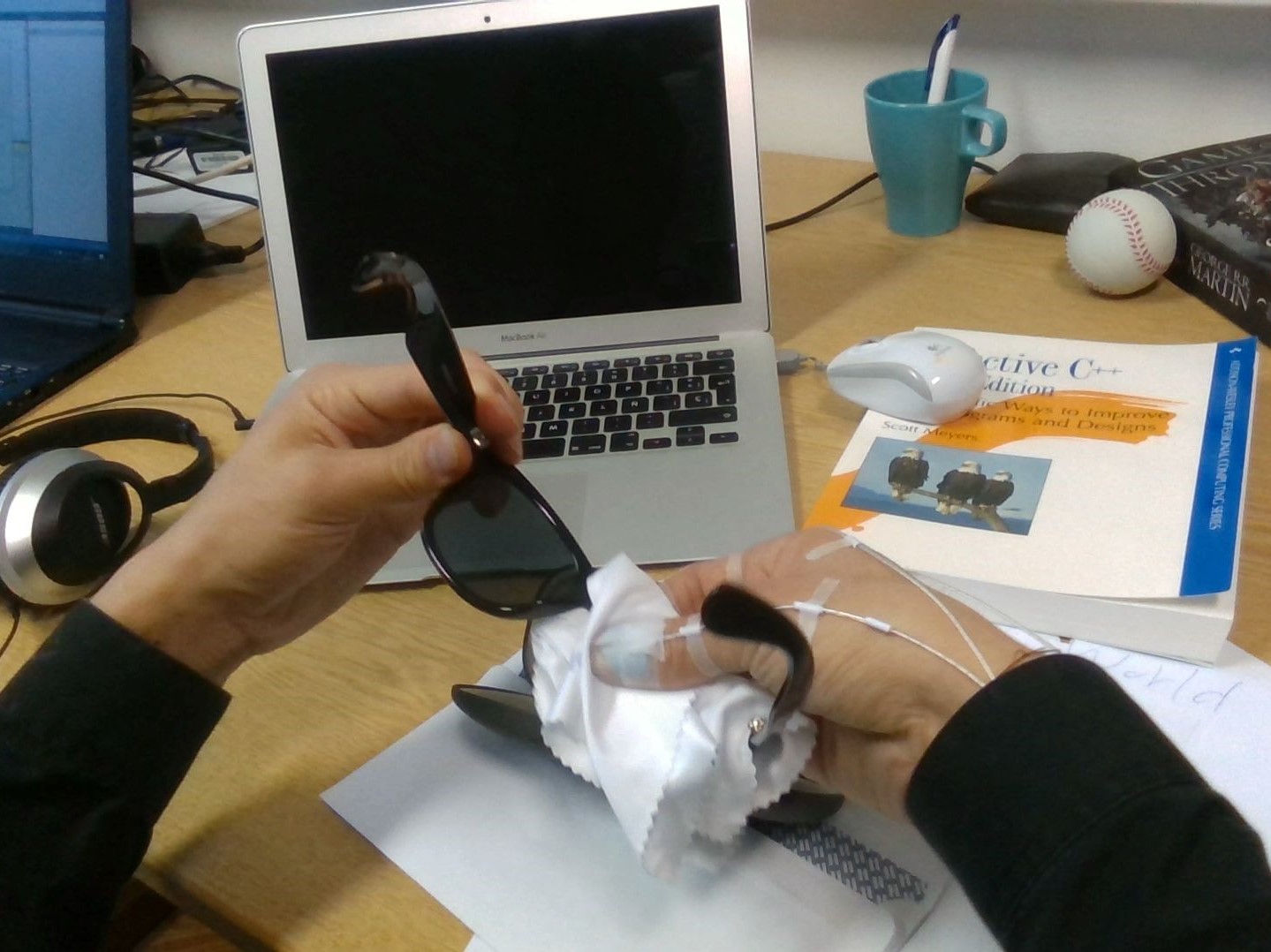}
    \includegraphics[width=0.3\linewidth, trim={0 15mm 0 0},clip]{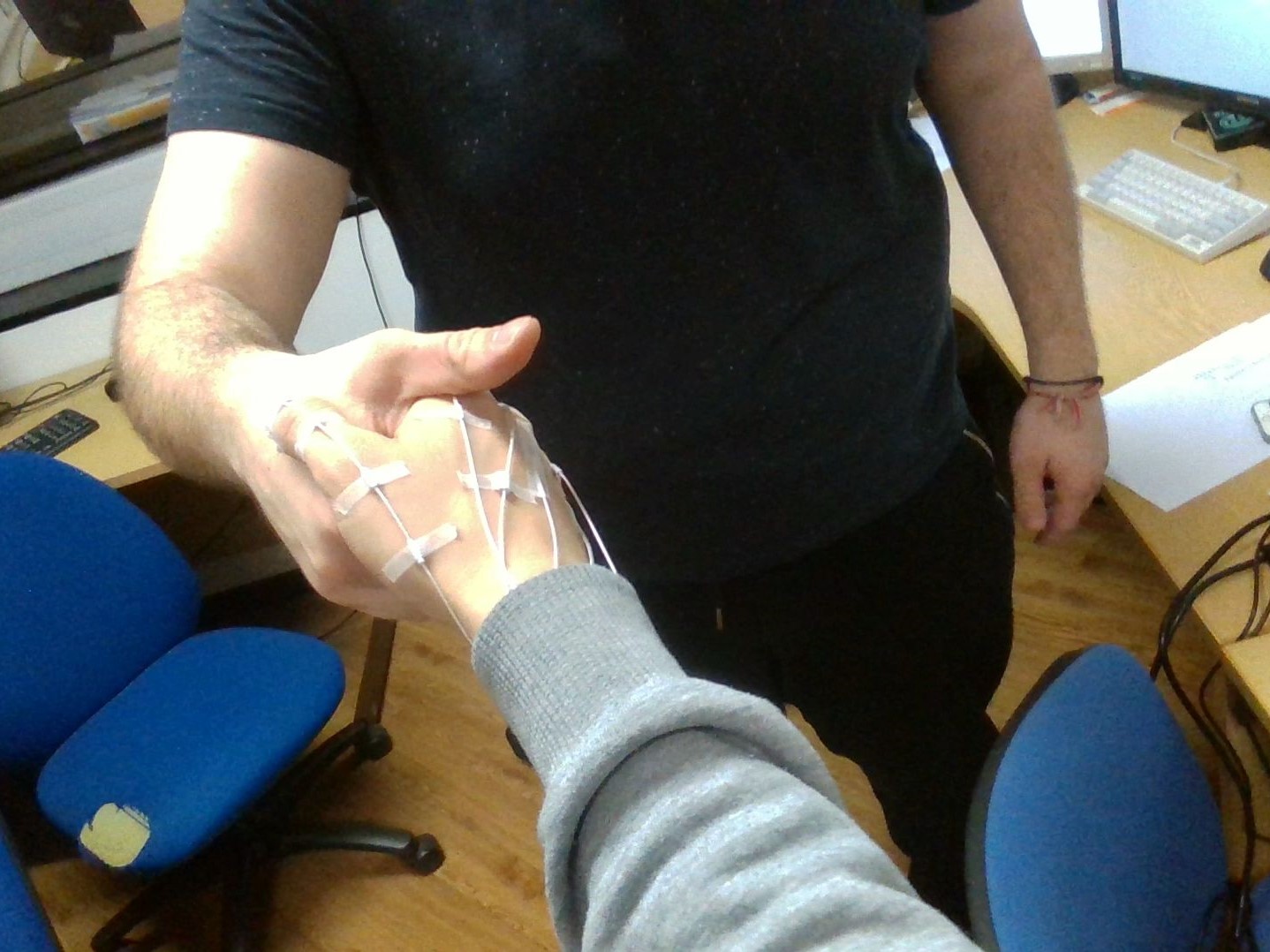}
    \includegraphics[width=0.3\linewidth, trim={0 15mm 0 0},clip]{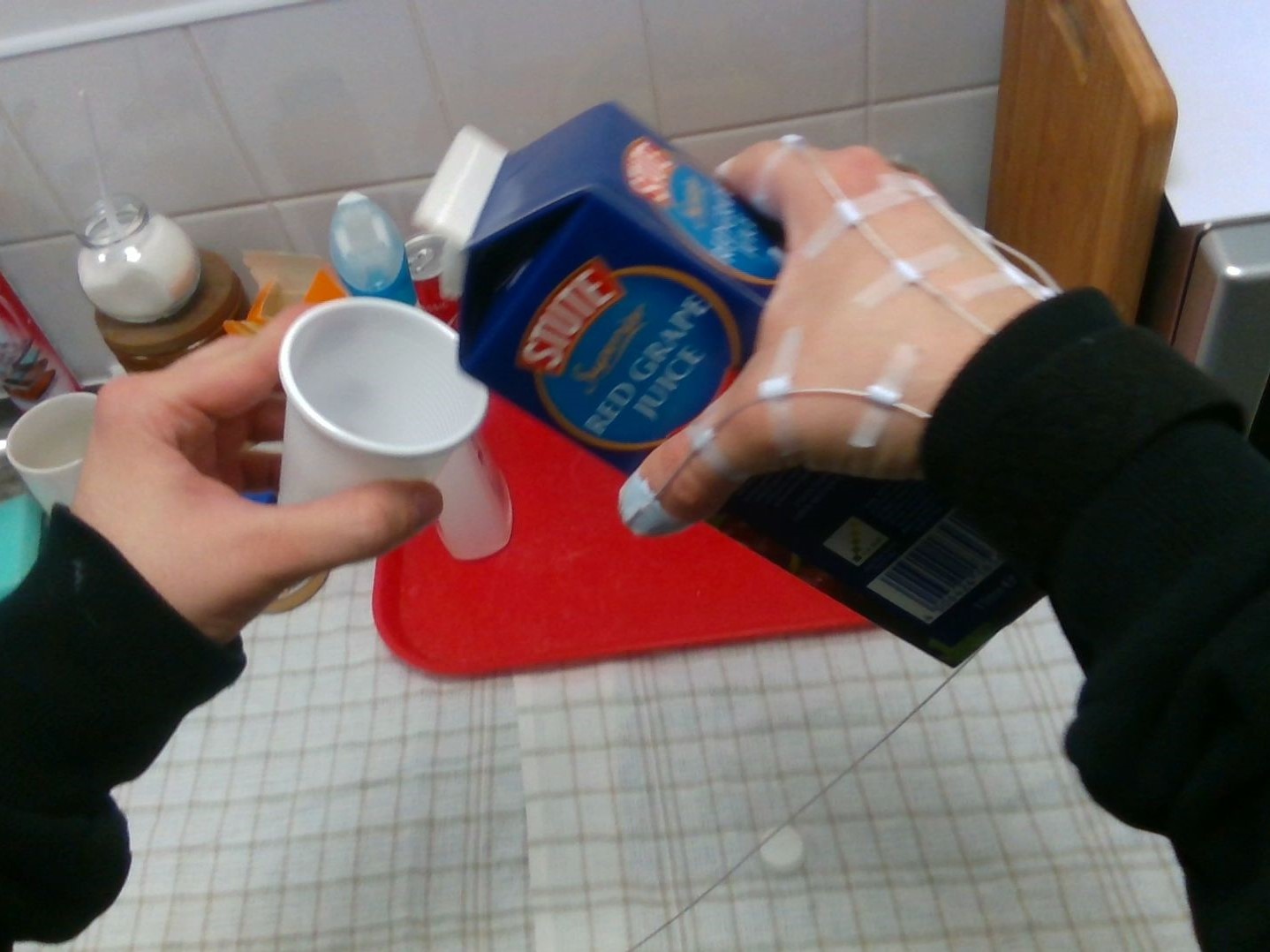}
    \caption{F-PHAB RGB sample frames}
\end{subfigure}
\begin{subfigure}[b]{\linewidth} 
\centering
    \includegraphics[width=0.3\linewidth]{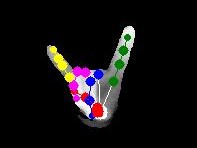}
    \includegraphics[width=0.3\linewidth]{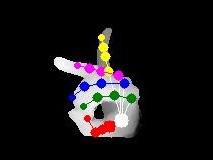}
    \includegraphics[width=0.3\linewidth]{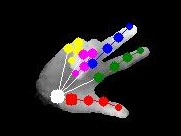}
    \caption{MSRA depth sample frames with skeleton joints}
\end{subfigure}
    \caption{Sample frames from the different evaluated data domains. (a) SHREC-17 dataset. Examples of actions \textit{grab}, \textit{expand} and \textit{rotation clockwise}. (b) F-PHAB dataset. Examples of actions \textit{clean glasses}, \textit{handshaking} and \textit{pour juice}. (c) MSRA dataset. Examples of the signs \textit{IP}, \textit{RP} and \textit{three}.}
    \label{fig:frame_samples}
\end{figure}

\subsubsection{Implementation and training details}
\paragraph*{Hand skeleton}
Since each dataset used provides different skeleton joints format, we use the 20 joints that SHREC-17 and F-PHAB have in common (see Fig. \ref{fig:min_hand}), and our proposed 7-joint skeletons representation, described in Section \ref{sec:hand_modeling}, suitable for the three datasets considered.

\paragraph*{Motion representation architecture}
our motion representation model backbone is a TCN with two stacks of residual blocks with dilations of 1, 2 and 4 for the layers within each block,
and convolutional filters of size 4, making a memory length of 32 frames long. Since the feature pre-processing filters out 2 out of 3 consecutive frames, this memory length covers 96 real frames.
Our backbone uses 256 filters in each convolutional layer, generating motion sequence descriptors of size 256. The summarization module reduces their dimensionality to 64 with a single 1D convolutional layer and then a single perceptron layer of size 32 generates the final descriptor weights. 
When the sequence summarization module is not used, the descriptor generated by the TCN at the last motion time-step is used for the action representation (\textit{Last TCN descriptor}). 
$\tau$ from Eq. \ref{eq:loss} is set as $0.07$.

\paragraph*{KNN classifier}
Our KNN classifier weights pairs of target-reference descriptors according to the inverse of their distance.
We validate the use of different number of neighbors,  i.e. 1, 3, 5, 7, 9, 11, and we report the results of the neighbor that optimizes the final classification accuracy. Additionally, the reference augmentation step increases the reference descriptors set randomly up to 40 times.

\subsection{Framework design evaluation}

This subsection analyzes and validates the main components of our framework using the \textbf{cross-domain approach} of Section \ref{sec:xdom}, since this setup is more demanding in terms of generalization capabilities.
We train our base motion representation model on the front view SHREC-17 dataset.
Then, we evaluate its accuracy on the egocentric F-PHAB validation splits (described in Section \ref{sec:datasets}).

To analyze the effect of different design choices, we start representing the motion sequences with the last descriptor generated by the TCN at the last time-steps (no use of the motion summarization module).

First, we show the benefits of using our proposed hand skeleton simplification. 
Table \ref{tab:skel_size} shows in each column the accuracy obtained in each of the F-PHAB validation splits.
Our proposed simplified 7-joint skeleton format reduces the coordinate redundancy and facilitates the generalization to other domains by reducing the overfitting on the source one. From now on, we set 7-joint skeleton format as default.

\begin{table}[!htb]
\centering
\begin{tabular}{|l|c|c|c|c|}
\hline
\textbf{Skeleton size} & \textbf{1:3} & \textbf{1:1} & \textbf{3:1} & \textbf{cross-person} \\ \hline
20 joints & 63.8 & 69.9 & 69.8 & 51.4 \\ \hline
7 joints & 66.3 & 71.0 & 73.8 & 53.5 \\ \hline
\end{tabular}
\caption{\textbf{Influence of the number of skeleton joints} in the hand representation. Motion representation model trained on SHREC. Action recognition accuracy validated on F-PHAB. }
\label{tab:skel_size}
\end{table}

Table \ref{tab:train_labels} shows the influence of using different classes to discriminate motion sequences while training the motion representation model.
Higher class granularity (28 action categories) manages to improve the cross-domain performance by learning more informative motion descriptors. From now on we set this class granularity as default for training.

\begin{table}[!htb]
\centering
\begin{tabular}{|c|c|c|c|c|}
\hline
\textbf{SHREC categories} & \textbf{1:3} & \textbf{1:1} & \textbf{3:1} & \textbf{cross-person} \\ \hline
14 & 58.3 & 65.4 & 65.9 & 48.9 \\ \hline
28 & 66.3 & 71.0 & 73.8 & 53.5 \\ \hline
\end{tabular}
\caption{\textbf{Influence of the training categories.} Motion representation model trained on SHREC with 7-joint skeletons. Action recognition accuracy evaluated on F-PHAB.}
\label{tab:train_labels}
\end{table}

Results from Table \ref{tab:summ_comp} show how our summarization module, from now on set as the default motion representation method, improves the classification accuracy with respect to the last descriptor of our TCN backbone. The summarization module suppresses noisy and non-informative per-frame descriptors, achieving a more informative motion representation. Interestingly, our motion summarization module learns good motion representations even when not many reference actions are available (splits 1:3). Another interesting finding is that augmenting the motion reference set helps to increase the accuracy in all the data splits by a noticeable margin.

However, we still find an accuracy drop when generalizing to actions of users not present in the motion reference set (cross-person splits). This is due to high inter-subject action variability of the F-PHAB dataset, and because no data from this dataset has been used to train our representation model.

\begin{table}[!htb]
\centering
\begin{tabular}{|l|c|c|c|c|}
\hline
\textbf{Action descriptor} & \textbf{1:3} & \textbf{1:1} & \textbf{3:1} & \textbf{cross-person} \\ \hline
\textbf{Last TCN descriptor} & 66.3 & 71.0 & 73.8 & 53.5 \\ \hline
\textbf{Summarization} & 70.6 & 75.5 & 77.7 & 58.4 \\ \hline
\textbf{Summarization*} & 76.2 & 79.7 & 82.0 & 62.7 \\ \hline
\multicolumn{5}{l}{\footnotesize \(^*\) includes an augmented motion reference set} \\
\end{tabular}
\caption{\textbf{Influence of the motion sequence summarization technique.} Motion representation model trained on SHREC (7-joint skeletons and 28 labels). Action recognition accuracy evaluated on F-PHAB. \textit{Last TCN descriptor}: descriptor generated by the TCN at the last time-step. \textit{Summarization}: descriptor generated by our summarization module.} 
\label{tab:summ_comp}
\end{table}

\begin{figure}[!htb]
    \centering
    \includegraphics[width=0.4\textwidth]{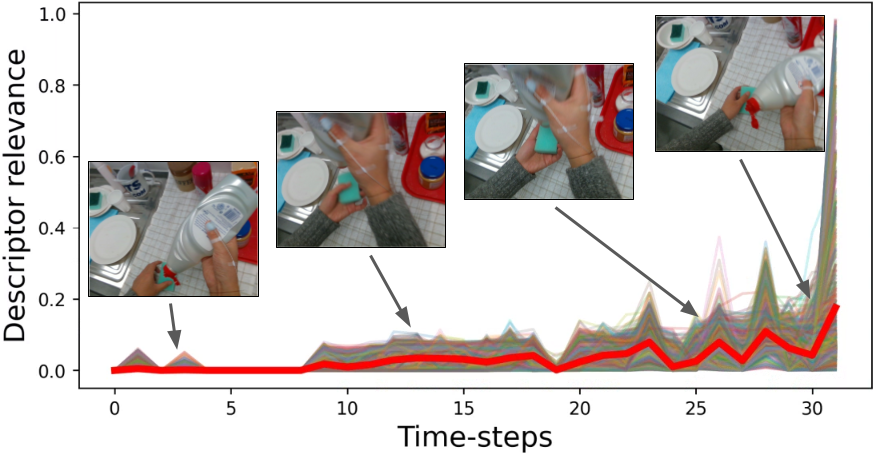}
    \caption{Relevance weights generated by the summarization module for each action sample from the F-PHAB validation split (1:1). Red line on top shows average of all generated weights. Frames correspond to the action of \textit{pour liquid soap}.
    }    
    \label{fig:att_weights}
\end{figure}

Figure \ref{fig:att_weights} shows the weights learned by our summarization module on the F-PHAB validation split (1:1). 
This plot illustrates the intuitive idea that later per-frame descriptors are more informative than earlier ones for final motion sequence representation.
However, computed weights do not exhibit a continuous growth along time, probably because contiguous time descriptors contain similar information. Although final descriptors may encode information about the whole action, they may also encode motion not related with the action itself but with idle poses for example. Therefore, they are not always the most relevant for the final action representation. 

\subsection{Cross-domain action classification}
\label{sec:fphab_bench}

This experiment evaluates the cross-domain generalization of our framework by classifying motion sequences from action categories and camera view-points not seen in the training data. 
For this experiment, we train our motion representation model as defined in Section \ref{sec:xdom} only on the front view SHREC-17 dataset (28 labels), and we evaluate it on the egocentric F-PHAB dataset. Results from our framework correspond to the processing of 7-joint skeletons and the use of our proposed motion summarization module.

\begin{table}[!tb]
\centering
\begin{tabular}{|l|c|c|c|c|}
\hline
\textbf{Model} & \textbf{1:3} & \textbf{1:1} & \textbf{3:1} & \textbf{cross-person} \\ \hline
\textbf{RGB \cite{feichtenhofer2016convolutional}} & -- & 75.3 & -- & -- \\ \hline
\textbf{Depth \cite{oreifej2013hon4d}} & -- & 70.61 & -- & -- \\ \hline
\textbf{LSTM \cite{zhu2016co}} & 58.75 & 78.73 & \textbf{84.82} & 62.06 \\ \hline
\textbf{DD-Net \cite{yang2019make}} & 75.09 & 81.56 & 88.26 & \textbf{71.8} \\ \hline
\textbf{Gram Matrix \cite{zhang2016efficient}} & -- & 85.39 & -- & -- \\ \hline
\textbf{Two-stream NN \cite{li2021two}} & -- & \textbf{90.26} & -- & -- \\ \hline\hline
\textbf{DD-Net \cite{yang2019make}} & 59.6 & 63.7 & 67.5 & 51.2 \\ \hline
\textbf{Ours} & 70.6 & 75.5 & 77.7 & 58.4 \\ \hline
\textbf{Ours*} & \textbf{76.2} & 79.7 & 82.0 & 62.7 \\ \hline
\multicolumn{5}{l}{\footnotesize \(^*\) includes an augmented motion reference set} \\
\end{tabular}
\caption{F-PHAB accuracy comparison. Upper block: results for methods trained on the F-PHAB dataset (\textit{intra-domain} classification). Bottom block: methods trained on SHREC-17 dataset (\textit{cross-domain} classification).}
\label{tab:phab_ablation}
\end{table}

Table \ref{tab:phab_ablation} shows the accuracy of the best performing methods on the F-PHAB dataset, trained as an intra-domain problem (upper block), and the results of our cross-domain approach (bottom block). The later include the evaluation of DD-Net \cite{yang2019make}, one of the best performing methods on the SHREC-17 classification benchmark. We used the available public code to train it with the SHREC-17 dataset (20-joint skeletons) as the authors state, extracting F-PHAB descriptors from its backbone and classifying them with our N-shot approach. 
Results from its evaluation show a lack of domain adaptation. Our method clearly outperforms the rest in this scenario.

The results show that our approach clearly outperforms the  RGB \cite{feichtenhofer2016convolutional} and depth-based \cite{oreifej2013hon4d} models trained on the target domain.
It is noticeable that we also get better or comparable results than a regular LSTM network \cite{zhu2016co} trained on the target dataset, specially when not many reference actions are available (1:3 split) or when not all the subjects are present in the reference split (cross-person splits). 
Although our cross-domain performance is behind the best intra-domain classification model \cite{zhang2016efficient}, we show later in Section \ref{sec:intra-dom-res} that we outperform them when training in the same domain. Remember that no specific training with the F-PHAB data splits has been performed in our evaluations.

\subsection{Cross-domain classification of long video sequences}\label{sec:online}
 
In this experiment we use the MSRA dataset, with hand motion sequences much longer than the memory of our representation model. 
This helps to illustrate two characteristics of our method. 
First, the motion summarization module (\ref{fig:pipeline}.c) not only helps to summarize the input motion, but also to enforce the TCN to generate informative per-frame descriptors (\ref{fig:pipeline}.b).
Second, per-frame descriptors can also be used to describe the input motion at each time-step and perform online and real-time recognition (see Section \ref{sec:time})

This experiment uses the same model trained in section \ref{sec:fphab_bench}. We evaluate its cross-domain performance in the MSRA dataset. Since sequences are too long for our summarization, we perform the KNN classification of all the motion descriptors generated by the TCN at each time-step (\ref{fig:pipeline}.b), denoted as \textit{online action classification}. We report the average of class probabilities of the frames within a video sequence for comparison with previous works, denoted as \textit{video classification}. For computational reasons, we randomly select just 8000 reference descriptors for the KNN evaluation.

Table \ref{tab:msra_bench} shows that, even though MSRA motion sequences do not correspond to the kind of motion seen in the training data, our approach achieves a high online per-frame classification.
Moreover, a simple average of the predicted frame probabilities results in a 97.1\% accuracy, comparable to current state-of-the-art results specifically trained on the MSRA dataset.
In this case, reference motion data augmentation does not provide an edge, probably because MSRA motion sequences already contain enough hand pose variations.

\begin{table}[!tb]
    \centering
\begin{tabular}{|l|c|c|}
\hline
\textbf{Model} & \textbf{Online classification} & \multicolumn{1}{l|}{\textbf{Video classification}} \\ \hline
\textbf{3D PostureNet \cite{liu20203d}} & -- & 98.56 \\ \hline
\textbf{Ours} & 85.8 & 97.1 \\ \hline
\textbf{Ours*} & 86.7 & 97.1 \\ \hline
\multicolumn{3}{l}{\footnotesize \(^*\) includes an augmented motion reference set} \\
\end{tabular}
    \caption{MSRA accuracy comparison. Batched vs. online predictions. Our results correspond to our motion representation model trained on SHREC-17 data (cross-domain).}
    \label{tab:msra_bench}
\end{table}

\subsection{Intra-domain classification and reference actions study}
\label{sec:intra-dom-res}
This experiment evaluates our method for intra-domain classification using the linear classifier from Section \ref{sec:intra_dom}. 

\subsubsection{SHREC-17 evaluation}

Table \ref{tab:shrec_bench} shows the classification accuracy of our framework trained and evaluated on the SHREC-17 dataset.
Results show that, even though our method was designed for cross-domain classification, it gets comparable results to the state-of-the-art when trained with the target dataset. Note that we are using just 7 out of the 22 original skeleton joints, that helps generalization to other datasets but it might lose domain-specific information.

\begin{table}[!hb]
\centering
\begin{tabular}{|l|c|c|c|}
\hline
\textbf{Model} & \textbf{SHREC 14} & \textbf{SHREC 28} \\ \hline
\textbf{DD-Net \cite{yang2019make}} & 94.6 & 91.9 \\ \hline
\textbf{Two-stream NN \cite{li2021two}} & 96.31 & 94.05 \\ \hline
\textbf{Ours} & 93.57 & 91.43 \\ \hline
\end{tabular}
\caption{Intra-domain classification on SHREC-17.}
\label{tab:shrec_bench}
\end{table}

\subsubsection{F-PHAB evaluation}

Table \ref{tab:fphab_bench} shows the classification accuracy of our framework trained and evaluated on each one of the F-PHAB data splits.
DD-net results, are obtained by training on the F-PHAB dataset with the original code and following the original paper \cite{yang2019make}.
Results show how we manage to outperform the current state-of-the-art in all the splits.
Interestingly, our method excels even when less training data is available (1:3). 
This generalization is visible even on the high inter-subject variability (cross-person) \cite{garcia2018first}.

\begin{table}[!htb]
\centering
\begin{tabular}{|l|c|c|c|c|}
\hline
\textbf{Model} & \textbf{1:3} & \textbf{1:1} & \textbf{3:1} & \textbf{cross-person} \\ \hline
\textbf{DD-Net \cite{yang2019make}} & 75.09 & 81.56 & 88.26 & 71.8 \\ \hline
\textbf{Two-stream NN \cite{li2021two}} & -- & 90.26 & -- & -- \\ \hline
\textbf{Ours} & \textbf{92.90} & \textbf{95.93} & \textbf{96.76} & \textbf{88.70} \\ \hline
\end{tabular}
\caption{Intra-domain classification on F-PHAB.}
\label{tab:fphab_bench}
\end{table}

\subsection{Time performance}\label{sec:time}
The presented work is a lightweight solution, able to perform online and real-time hand action recognition (like in Section \ref{sec:online}).
Our base motion representation model (\ref{fig:pipeline}.b) gets per-frame descriptors in 0.8 ms per time-step in GPU (NVIDIA GeForce GTX 1070) 
and 1 ms in CPU (Intel Core  i7-6700).
The motion summarization module (Fig. \ref{fig:pipeline}.c) and the linear classifier (intra-domain) from Section \ref{sec:intra_dom} can be used at a negligible cost.
The KNN classifier (cross-domain) from Section \ref{sec:xdom} has a cost \(O(k * log(n))\) that depends on the number of neighbors \(k\) and the size \(n\) of the motion reference set. 
For instance, the KNN classification on the 1:1 data split from F-PHAB (\(575\) motion reference sequences), just takes \(0.2\) ms per motion descriptor when using \(5\) neighbors and \(0.5\) ms when augmenting the motion reference set \(40\) times.

\section{Conclusions}

The present work introduces a hand action recognition solution, specifically designed to be robust to different action domains and camera perspectives, and able to perform in online and real-time domains.
Our framework extracts, from skeleton motion sequences, sets of pose features adapted to heterogeneous motion kinematics.
Then, our motion representation model uses a Temporal Convolutional Network that generates per-frame motion descriptors, and a simple motion summarization module weights them, according to their relevance, generating the final motion representation.
We trained and validated our motion representation model in two different conditions. In intra-domain classification, we achieve better or similar results than state of the art methods 
in well-known benchmarks.
More importantly, in cross-domain classification, our approach is able to generalize to unseen target action domains and camera view-points, achieving comparable results to the state-of-the-art methods trained on the target data domains.

{
\bibliographystyle{IEEEtran}
\bibliography{biblio}
}

\end{document}